\documentclass{article}
\usepackage{spconf}
\usepackage{blindtext}
\usepackage{amssymb}
\usepackage{amsmath}
\usepackage{fancyhdr}
\usepackage{graphicx}
\usepackage{float}
\usepackage{algorithm}
\usepackage{algpseudocode}
\usepackage{subcaption}
\usepackage{multirow}
\usepackage[algo2e]{algorithm2e}
\usepackage{color}
\usepackage{url}

\usepackage[colorlinks=true, linkcolor=blue, citecolor=blue, urlcolor=black]{hyperref}

\usepackage{bm}

\newfont{\bbb}{msbm10 scaled 700}

\newcommand{\cv}{{\bf c}}

\newcommand{\fv}{{\bf f}}

\newcommand{\mv}{{\bf m}}

\newcommand{\pv}{{\bf p}}
\newcommand{\qv}{{\bf q}}

\newcommand{\tv}{{\bf t}}

\newcommand{\vv}{{\bf v}}
\newcommand{\xv}{{\bf x}}
\newcommand{\yv}{{\bf y}}

\newcommand{\Cm}{{\bf C}}
\newcommand{\Dm}{{\bf D}}

\newcommand{\Id}{{\bf I}}

\newcommand{\Lm}{{\bf L}}

\newcommand{\Um}{{\bf U}}
\newcommand{\Wm}{{\bf W}}
\newcommand{\Vm}{{\bf V}}

\newcommand{\Ym}{{\bf Y}}

\newcommand{\Nc}{{\cal N}}

\newcommand{\Lambdam}{\hbox{\boldmath$\Lambda$}}
\newcommand{\Deltam}{\hbox{\boldmath$\Delta$}}

\renewcommand{\arg}{{\hbox{arg}}}

\begin{document}
\title{Motion estimation and filtered prediction for dynamic point cloud attribute compression}
%
\name{Haoran Hong$^{\star}$, Eduardo Pavez$^{\star}$, Antonio Ortega$^{\star}$, Ryosuke Watanabe$^{\star,\dagger}$, Keisuke Nonaka$^{\dagger}$
\thanks{This work was funded in part by KDDI Research, Inc.~and by the National Science Foundation (NSF CNS-1956190).}}
\address{$^{\star}$University of Southern California, Los Angeles CA. 90089 USA \\
$^{\dagger}$KDDI Research, Inc., Japan\\}
\maketitle
%
%
%

\ninept

\begin{abstract}
In point cloud compression, exploiting temporal redundancy for inter predictive coding is challenging because of the irregular geometry. This paper proposes an efficient block-based inter-coding scheme for color attribute compression. The scheme includes integer-precision motion estimation and an adaptive graph based in-loop filtering scheme for improved attribute prediction. The proposed block-based motion estimation scheme consists of an initial motion search that exploits geometric and color attributes, followed by a motion refinement that only minimizes color prediction error. To further improve color prediction, we propose a  vertex-domain low-pass graph filtering scheme that can adaptively remove noise from predictors computed from motion estimation with different accuracy. Our experiments demonstrate significant coding gain over state-of-the-art coding methods. 
\end{abstract}
\begin{keywords}
inter prediction, geometry-based point cloud coding, motion estimation, low-pass filter, graph filter.
\end{keywords}

\section{Introduction}
In current video coding systems, motion estimation (ME) and motion compensated prediction (MCP) are two essential techniques to exploit temporal redundancy between frames, so that inter-predictive coding techniques can be applied. 
In video-based point cloud  compression (V-PCC) \cite{overview2020} a direct extension of these motion-based techniques to dynamic point cloud compression (PCC) is straightforward. 
However, for geometry-based PCC  systems, where a point cloud (PC) is encoded directly in 3D space, ME becomes challenging. 
To see why, note that within a block in a video frame \textit{all} pixel positions contain color information. Thus, both the reference and predictor blocks have the same number of pixels. 
In contrast, since PCs represent the surface of objects within a volume. Each 3D block in general has \textit{a different geometry and a different number of occupied voxels}. Due to this irregular spatial structure, ME for geometry-based PC coding has to rely on PC registration (PCR) techniques \cite{PCRsurvey2021}, which match point sets of different sizes and geometries. 

While PCR techniques have been widely studied and play a critical role in many computer vision applications, they need to be modified for ME and MCP, where the goal is to achieve better temporal prediction of PC attributes such as color. 
ME can be viewed as the estimation of a 3D transformation between two  PCs, which seeks to align each element in a set of non-overlapping units  in the predicted PC (e.g., voxels, blocks or clusters) to a corresponding unit in the reference PC. 
As an example, each block in the predicted PC is matched to a corresponding block in the reference and the transformation that best aligns the blocks is the estimated motion. 
Existing geometry-based ME techniques are based on  
 iterative closest point (ICP) \cite{Rufael2017,Xu2020,SantosICIP2021}, 
 graph matching  \cite{Dorina2016} and block matching  \cite{Ricardo2017,Camilo2018,Camilo2019}.

ICP-based methods are the most commonly used due to their good performance and efficiency. They work particularly well when the two PCs to be matched are nearly aligned to begin with, which is usually the case for consecutive frames in  dynamic PCs  captured at high frame rates (e.g., $30$ frames per second).  
In contrast, other PCR methods for ME are more complex and cannot be easily scaled to large PCs. Thus, our proposed ME is based on ICP.

MCP for geometry-based inter predictive coding can be accomplished by simply replacing the voxels in a predicted block with voxels in its best matching reference block  \cite{Rufael2017,Ricardo2017}, with mode decision determining whether intra coding should be used instead if a sufficiently good approximation is not given by the copied block.  
As an 
alternative, in \cite{Xu2020,Souto2020} the color  of each voxel in the current frame is predicted 
by the color of its corresponding voxel in the reference frame and a residual signal is sent. 
Our proposed method uses blocks instead of clusters as the basic unit for ME (as \cite{Rufael2017,Ricardo2017}) but generates residues (as \cite{Dorina2016,pavez2018dynamic,Xu2020}). In addition, we 
propose a vertex-domain low pass graph filter that leads to significant improvements in prediction.

Our work addresses several limitations in recent work. 
First, existing block-based ME schemes \cite{Rufael2017,Ricardo2017,Camilo2018,Camilo2019,SantosICIP2021}  mimic video coding approaches  by matching blocks of \textit{same size} in the predicted and reference frames. 
Instead, as in \cite{Xu2020} we directly optimize the alignment of a block in the current frame to a \textit{larger} bounding box in the reference frame. With this approach, the closest points matched in the reference frame are not restricted to be within a block of same size as the block in the current frame. 
Note that using blocks is more efficient than using clusters, as in \cite{Xu2020}, since no clustering operations are needed, and blocks can be efficiently obtained using the octree structure. Also, different from recent ME schemes \cite{Xu2020,Rufael2017,SantosICIP2021} we use a modified color-ICP \cite{colorICP1999} without rotation transformation (to avoid sending rotation information as overhead), with a hybrid matching metric that combines geometry and color, which leads to better inter prediction performance. 

Even though our proposed block-based ME is based on a criterion that combines  color and geometry (using color-ICP \cite{colorICP1999}) there is no guarantee that it is optimal for predicting  PC color attributes. 
Thus, after the initial block matching, we propose two novel techniques not considered in the geometric point cloud coding literature.  
First, we use 
a \textit{refined local search}  around the color-ICP based motion to generate a refined motion  that minimizes the color prediction error. By minimizing residual energy we improve prediction, while preserving geometric consistency  because we only search locally around the initial motion found through color-ICP. 
Second, for MCP, we propose a vertex-domain low-pass graph filter. Unlike previous approaches \cite{Xu2020}, the proposed adaptive filtering scheme has low computational complexity since it  can be implemented with sparse matrix vector products. The proposed   adaptive filtering scheme can  eliminate high frequency components in predictors and therefore improve  prediction when there is inaccurate motion.
Our method can outperform transform coding baselines,  \cite{RAGFT} and \cite{raht} , with average gains of $3.21dB$ and $4.95dB$, respectively, and outperform the state-of-the-art inter coding scheme \cite{Xu2020} with average gains of $2.15dB$.


\section{Block-based motion estimation}
\label{sec:ME}
\subsection{Notation}
A 3D point cloud $P$ consists of a list of points in $\mathbb{R}^3$ with their coordinates, along with a list of point attributes, such as color. Each point/attribute can be denoted by $\pv = (\vv,\cv)$,  where $\vv = (x,y,z)$ are the Cartesian coordinates,  and  $\cv = (Y,U,V)$ are their corresponding color attributes in YUV format.  A  dynamic PC is a sequence of PCs denoted by $\lbrace P_{1}, \cdots, P_{T}\rbrace$. We also denote by $\Vm_t $ and $\Cm_t$, the $N_t \times 3$ matrices containing the point coordinates and color attributes at time $t$, respectively.
\subsection{Motion compensated compression of dynamic point clouds}
\label{ssec:mc_prediction}
MC prediction for inter coding of 3D PC color attributes has been proposed with various types of motion vectors \cite{Xu2020,Souto2020,FME2022}.  There is a fundamental difference between MC prediction in video and 3D PCs. Because predicted and reference video blocks have the same number of pixels and geometry, there is a $1$ to $1$ mapping between reference and predicted pixels. In contrast, because 3D PC geometry is irregular, and blocks may have different number of points, such correspondences are not readily available.
We next describe the process of motion compensated matching between the target and reference blocks in more detail. 

Let $P_{t,i}  \subset P_t$ be the $i$th block  of a PC at time $t$, with points $\Vm_{t,i}$ and colors $\Cm_{t,i}$, and  let $\mv_{t,i}$ be its corresponding motion vector. 
To encode the color attributes  $\Cm_{t,i}$ of the $i$th block, given the motion vector $\mv_{t,i}$ and the decoded attributes $\hat{\Cm}_{t-1}$ of $P_{t-1}$, we define a predictor  by finding for each $\pv_j=[\vv_j, \cv_j] \in P_{t,i}$ its  motion-compensated nearest neighbor in the reference frame: 
\begin{equation}\label{eq:MCpredict}
 \tilde{\pv}_j=[\tilde{\vv}_j, \tilde{\cv}_j] = \arg\min_{ \qv = [\vv,\hat{\cv}] \in P_{t-1}} \Vert \vv_j + \mv_{t,i} -  \vv\Vert^2.
\end{equation}
Note that given $\mv_{t,i}$, which is transmitted as overhead, the decoder can use \eqref{eq:MCpredict} to find the nearest neighbor in $P_{t-1}$ of any point in $P_{t,i}$ 
(PC geometry for $P_t$ is encoded without loss and available at the decoder). 
The predictor for $\Cm_{t,i}$ is the matrix $\tilde{\Cm}_{t,i} = [\tilde{\cv}_j]$. 
The residual $\Cm_{t,i} - \tilde{\Cm}_{t,i}$ can then be transformed,  quantized,  and entropy coded. 
In \autoref{sec:ICPmotion} and \autoref{sec:RFmotion} we describe our proposed ME algorithm to generate $\mv_{t,i}$. Later in \autoref{sec:LP_filter} we introduce an in-loop filter for enhancing the MC predictor $\tilde{\Cm}_{t,i}$.
\subsection{Block based color-ICP for initial ME}
\label{sec:ICPmotion}
The current frame $P_t$  is partitioned into non-overlapping blocks (e.g., using the octree structure). The $i$th block is denoted by  $P_{t,i}\subseteq P_t$, as in \autoref{ssec:mc_prediction}. 
ICP algorithms iterate two steps until convergence: 1) point matching, and 2) transformation. Because our goal is color compression, we consider the hybrid metric between points with undistorted original colors $\pv_1=[\vv_1, \cv_1]$, and  $\pv_2=[\vv_2, \cv_2]$, given by
\begin{align}
\delta(\pv_1,\pv_2) &= \alpha\Vert \vv_1 - \vv_2\Vert^2 + (1-\alpha)\Vert \cv_1 - \cv_2\Vert^2,
\label{equ:NNcriterion}
\end{align}
where $\alpha \in [0,1]$, so  that  both geometry and  color are used for  matching. While in traditional ICP, transformations consist of both rotation and translations,  we only allow translations in integer precision as part on our transformations,  to reduce the overhead of transmitting motion vectors. This leads to the following iterative steps. 

\textbf{Matching:} For each $\pv_j \in P_{t,i}$, we find the best matching point in the reference frame by solving
\begin{equation}\label{eq:ICP:matching}
    \pv'_j=[\vv'_j, \cv'_j] = \arg\min_{\qv \in P_{t-1}} \delta(\pv_j, \qv),
\end{equation}
where $\delta$ is defined in \eqref{equ:NNcriterion}.

\textbf{Translation:}
The optimal translation $\tv_{t,i}$ is the one that minimizes the average geometric distance between the points in $P_{t,i}$, and their matches in the reference, thus
\begin{equation}
\tv_{t,i} = \arg\min_{\tv}  \sum_{j: \pv_j \in P_{t,i}}{\Vert \vv_{j} + \tv - \vv'_{j} \Vert^2},
\label{equ:IME}
\end{equation}
where $\vv'_j$ is computed using \eqref{eq:ICP:matching}.
It can be shown that $\tv_i = ({1}/{\vert P_{t,i}\vert})  \sum_{j=1}^{|P_{t,i}|} (\vv'_j - \vv_j)$ \cite{ICP1992}.
After the optimal translation is found with \eqref{equ:IME},   for each point $\pv_j = [\vv_j, \cv_j] \in P_{t,i}$, we shift its coordinates   towards the reference, that is, $\vv_j \gets \vv_j + \tv_{t,i}$, and then the \textbf{matching} and \textbf{translation} steps are repeated until convergence.  Note that at the end of the $\ell$th iteration, the translation step outputs a translation vector  which we store and denote by $\tv_{t,i}^{\ell}$. 
%
%
The final color ICP  motion for each block is computed by adding $\tv_{t,i}^{\ell}$ over all iterations and rounding up to integer precision resulting in
\begin{equation}
\tv^{ICP}_{t,i} = \text{round}\left(\sum_{\ell \geq  1} {\tv^{\ell}_{t,i}} \right).
\label{equ:ICPmotion}
\end{equation}

\subsection{Locally refined motion search}
\label{sec:RFmotion}
Even if the color-ICP motion vector preserves original color and geometric consistency, it is not guaranteed to be optimal for predicting (and coding) color attributes, especially since the reconstructed reference is distorted by lossy coding.
We propose to refine the motion vector based on distorted reconstructed colors to further minimize the prediction error of the MC predictor from \autoref{ssec:mc_prediction}. 
For each block $P_{t,i}$, we take $\tv^{ICP}_{t,i}$ as the initial motion and apply exhaustive search within a small search range to locally refine it.
The coordinates of  candidate displacement vectors $\Deltam$ are restricted to be integers in $\left[ -\beta, \beta \right]$, resulting in $(2\beta+1)^3$ possible MVs. For each point $\pv_i \in P_{t,i}$ in the current block and candidate motion vector $\tv^{ICP}_{t,i}+\Deltam$ we compute the MC predictor using \eqref{eq:MCpredict} resulting in $\tilde{\cv}_j(\tv^{ICP}_{t,i}+\Deltam)$. 
Then the optimal displacement is the one that minimizes color distance for all voxels in the predicted  block, 
\begin{equation}
    \Deltam_{t,i} = \arg\min_{\Deltam}\sum_{j: \pv_j \in P_{t,i}} \Vert \cv_j -  \tilde{\cv}_j(\tv^{ICP}_{t,i}+\Deltam) \Vert^2.
\end{equation}
The optimal motion vector for  block $P_{t,i}$ is  
\begin{equation}
\mv_{t,i} = \tv^{ICP}_{t,i} + \Deltam_{t,i}.
\label{equ:optIMV}
\end{equation}
\section{In-loop graph filtering}
\label{sec:LP_filter}
Following \autoref{ssec:mc_prediction}, for a given   motion vector $\mv_{t,i}$ for the $i$th block, we can obtain the MC predictor for the block attributes  $\tilde{\Cm}_{t,i}$, which can be computed using \eqref{eq:MCpredict}. In video coding, it is well known that significant coding gains can be obtained by removing noise from predictors with some form of in-loop filtering. 
Since PCs have an irregular spatial structure, we propose to use graph filters for this purpose (refer to \cite{gene2018} for an overview of graph filters for image processing).
We now describe a novel in-loop graph filtering technique to refine color attribute prediction.
\subsection{Graph based in-loop filtering}
A graph $G$ is defined as $G=(V,E)$, where $V$ is the set of vertices and $E\subseteq V\times V$ is the set of edges. The $a$-th vertex is denoted  $v_a\in V$ and the edge connecting $v_a$ and $v_b$ is denoted as $e_{ab}\in E$. 
The weight of an edge is denoted as $W_{a,b}$. 
We can represent the graph connectivity information using a $n\times n$ \textit{adjacency matrix} $\Wm$, which has zero diagonal elements and $\Wm_{a,b} = W_{a,b}$ for off-diagonal terms. A \textit{degree matrix} $\Dm$ is a $n\times n$ diagonal matrix with $D_{a,a} = \sum_b W_{a,b}$.
A \textit{combinatorial Laplacian matrix} is defined as $\Lm = \Dm-\Wm$. 
A \textit{random walk Laplacian matrix} is defined as $\Lm_{RW} = \Dm^{-1}\Lm$, which can be eigen-decomposed as $\Lm_{RW} = \Um \Lambdam \Um^{-1}$ \cite{ortega2022introduction,IRAGFT2018}.

For 3D PCs, we construct a graph with random walk Laplacian $\Lm^{(t,i)}_{RW}$ for each block $P_{t,i}$, where the edge weights $W_{a,b}$ are inversely proportional to the geometric distance $\Vert \vv_a - \vv_b\Vert^2$ (see experiments for more details), while  the color signal $\Cm_{t,i}$, and its MC predictor $\tilde{\Cm}_{t,i}$ are the graph signals.  
\subsection{Proposed graph filter}
\label{sec:LP}

Let $\xv$ and ${\yv}$ be the graph signals before and after filtering, 
respectively. We define a one-hop weighted average filter as
\begin{equation}
\label{equ:nodedomain}
y(a) = \frac{1}{2D_{a,a}}(D_{a,a}x(a) +  \sum_{b \in \Nc(a,1)} W_{a,b} x(b)), 
\end{equation}
which can be written in vector form as: 
\begin{equation}
    \yv = \left( \Id - \frac{1}{2}\Lm_{RW} \right) \xv = h(\Lm_{RW}) \xv,
\end{equation}
where $h (\lambda) = 1- \frac{1}{2}\lambda$ and $h(\Lm_{RW}) = \Um h(\Lambdam)U^{-1}$.
Since \eqref{equ:nodedomain} can be computed locally when the graph is sparse,  this filter can be computed efficiently. Because $\lambda \in [0,2]$, the filter $ h(\lambda)$ takes values in  $[0,1]$ thus acting as a low-pass filter that preserves the DC component, i.e., $\tilde{\fv} = \mathbf{1}$ whenever $\fv =\mathbf{1}$. 
For the $i$th block in the PC at time $t$, we propose using the filtered predictor
\begin{equation}\label{eq:predictor_filtered1}
    \tilde{\Cm}^{(1)}_{t,i} = h(\Lm_{RW}^{({t,i})}) \tilde{\Cm}_{t,i}.
\end{equation}
\subsection{Adaptive filtering for improved prediction}
\label{sec:adaptiveLP}
\begin{figure}[ht]
\begin{center}
\includegraphics[width = 0.25\textwidth]{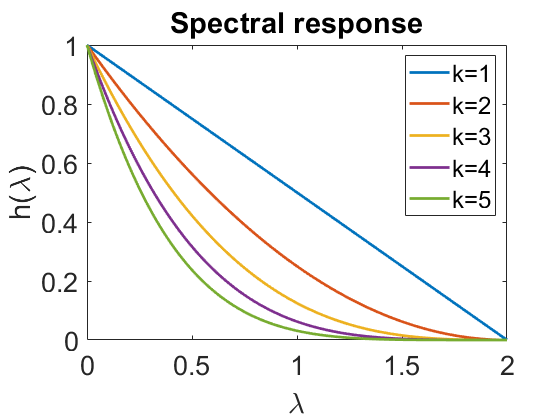}
\end{center}
\vspace{-5mm}
\caption{Spectral response of the proposed filter.}
\label{fig:specres_proposed}
\vspace{-4mm}
\end{figure}

Since motion estimation accuracy, compression error and noise may vary from block to block, we consider a block-level adaptive filtering approach where the filter from \eqref{eq:predictor_filtered1} is applied  $k$ times, resulting in the refined predictor
\begin{equation}
\label{equ:K_LP}
\tilde{\Cm}^{(k)}_{t,i} = h^k(\Lm_{RW}^{({t,i})}) \tilde{\Cm}_{t,i}.
\end{equation}
As  $k$ increases the filter \eqref{equ:K_LP} has a sharper frequency response, as shown in  \autoref{fig:specres_proposed}, thus providing additional flexibility for removing high frequency noise. 
For each  block, we choose $k$ to minimize the color error, that is, 
\begin{equation}
k_{t,i}^*  = \arg\min_{k \geq 0} \Vert \Cm_{t,i} - \tilde{\Cm}^{(k)}_{t,i} \Vert^2.
\label{equ:optimal_k}
\end{equation}
When $k_{t,i}^*=0$, we recover the   MC predictor from  \autoref{ssec:mc_prediction}. The sequence of $k_{t,i}^*$ for all blocks is encoded and sent as overhead.
%

\section{Experiments}
\label{sec:experiments}
\subsection{Dataset and parameter settings}
We evaluate the proposed integer motion estimation scheme for compression of color attributes of the \texttt{8iVFBv2} dataset \cite{8idataset}, which consists of four sequences: \texttt{longdress}, \texttt{redandblack}, \texttt{loot} and \texttt{soldier}. We  use the first $8$ frames of  each sequence.
%
We implement a conventional inter-coding system, where previously decoded frames are used as reference. The group of frames (GOF) is set to be 8, wherein the first frame is coded in intra mode, and the rest are coded in inter mode. The prediction and transform units are both $16\times16\times16$. We use block graph Fourier transform (block-GFT) \cite{GFT2014} to transform residues and choose $\alpha=0.1$\footnote{Performance was robust to a very wide range of $\alpha$ and very minor changes were observed in the results with $\alpha\in [0.1,0.5]$.} in \eqref{equ:NNcriterion}. 
To reduce the complexity of nearest point search in \autoref{sec:ICPmotion} we match each predicted block to a larger region (a block of size $61\times61\times61$) centered around the zero motion point in the reference frame.
In \autoref{sec:RFmotion} we set $\beta=1$, so that $27$ candidate motion vectors are evaluated during the refinement stage. For filtering, we construct a threshold graph with weight equal to the reciprocal of the geometric distance, and  voxels within a block are connected to each other if their Euclidean distance is less than or equal to $1$. The constructed graph is also used for block-GFT.
In  \eqref{equ:optimal_k} we choose integer values $k\in [ 0, 5]$.  For all schemes (proposed and baselines) the quantized coefficients are entropy coded by the adaptive run-length Golomb-Rice algorithm (RLGR) \cite{RLGR} while the overheads of $m_{t,i}$s and $k^*_{t,i}$s computed for blocks are separately entropy coded 
with the Lempel–Ziv–Markov chain algorithm of \cite{LZMA}. Other experimental settings are the same as in \cite{FME2022}.
%
Rate is measured in bits per voxel (bpv), while decoded quality is measured in average peak signal-to-noise ratio over Y component (PSNR-Y), 
%
\begin{equation*}
PSNR_Y = -10 \log_{10} \left (\frac{1}{T} \sum_{t = 1}^T \frac{ \| \Ym_t - \hat{\Ym}_t  \|_2^2 }{ 255^2  N_t} \right),  bpv = \frac{\sum_{t = 1}^T b_t}{\sum_{t = 1}^T N_t},
\label{equ:psnrY}
\end{equation*}
where $\Ym_t$ and $\hat{\Ym}_t$ represent original and reconstructed luma signals   of the $t$-th frame, respectively, and $T$ is the total number of frames.  $b_t$ is the total bitrate required to encode YUV components and overhead (when necessary) of $t$-th frame.  $N_t$ is the total number of occupied voxels in $t$-th frame.
%

\subsection{Analysis  of the proposed schemes} 
We evaluate our proposed ME and adaptive filtering method 
and show how each component contributes to the overall inter coding performance. We considered the following schemes: ME using geometric ICP and color-ICP of the hybrid criterion are denoted as \texttt{Geo} and \texttt{Geo+Col}, where added local refinement search is denoted as \texttt{Geo+Col+RF}. All the above schemes do not use any filtering scheme. Then adding a low pass filtering scheme with fixed $k = 1$ is denoted by \texttt{Geo+Col+RF+LP}, while the version with adaptive filtering is denoted \texttt{Geo+Col+RF+LP(k)}. 
Except \texttt{Geo}, all the above schemes that integrate color in ME, include coded MVs in the overhead. For \texttt{Geo+Col+RF+LP(k)}, the optimal $k^*_i$ chosen for each block is coded as overhead.


\begin{figure}[htb]
\begin{subfigure}[t]{0.25\textwidth}
\includegraphics[width = \linewidth]{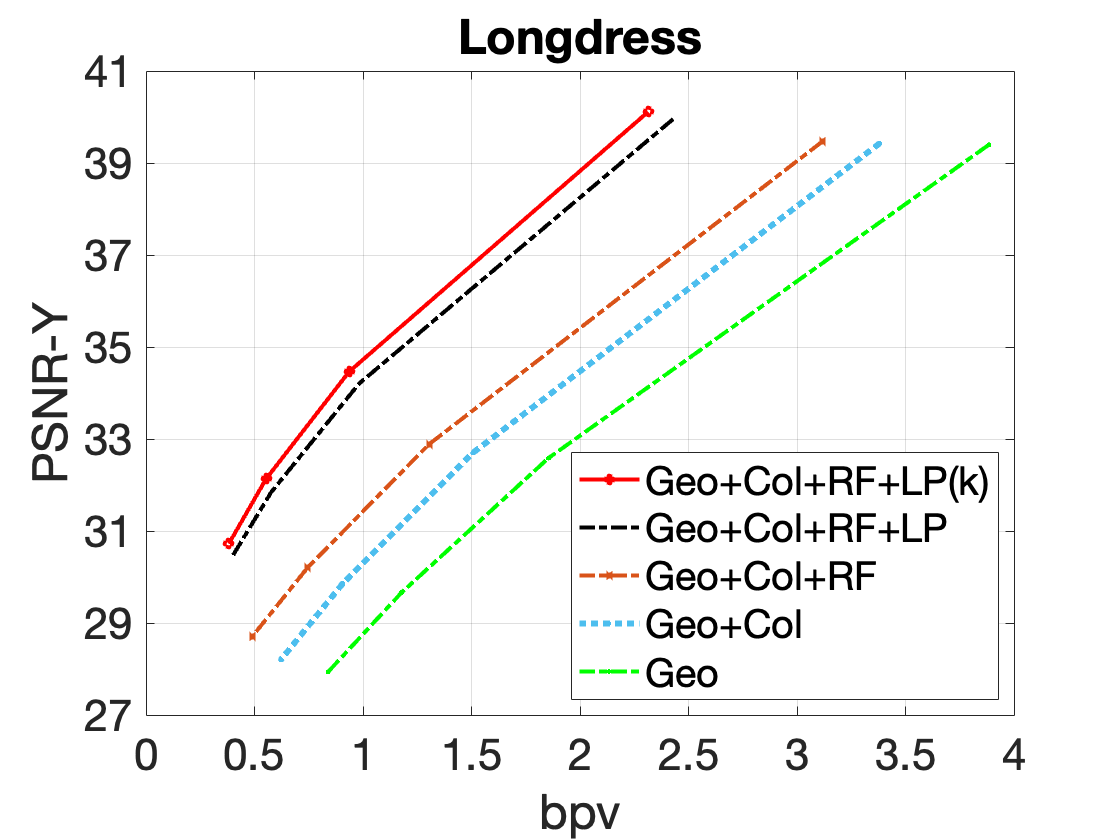}
\end{subfigure}
\hspace{-5.5mm}
\begin{subfigure}[t]{0.25\textwidth}
\includegraphics[width = \linewidth]{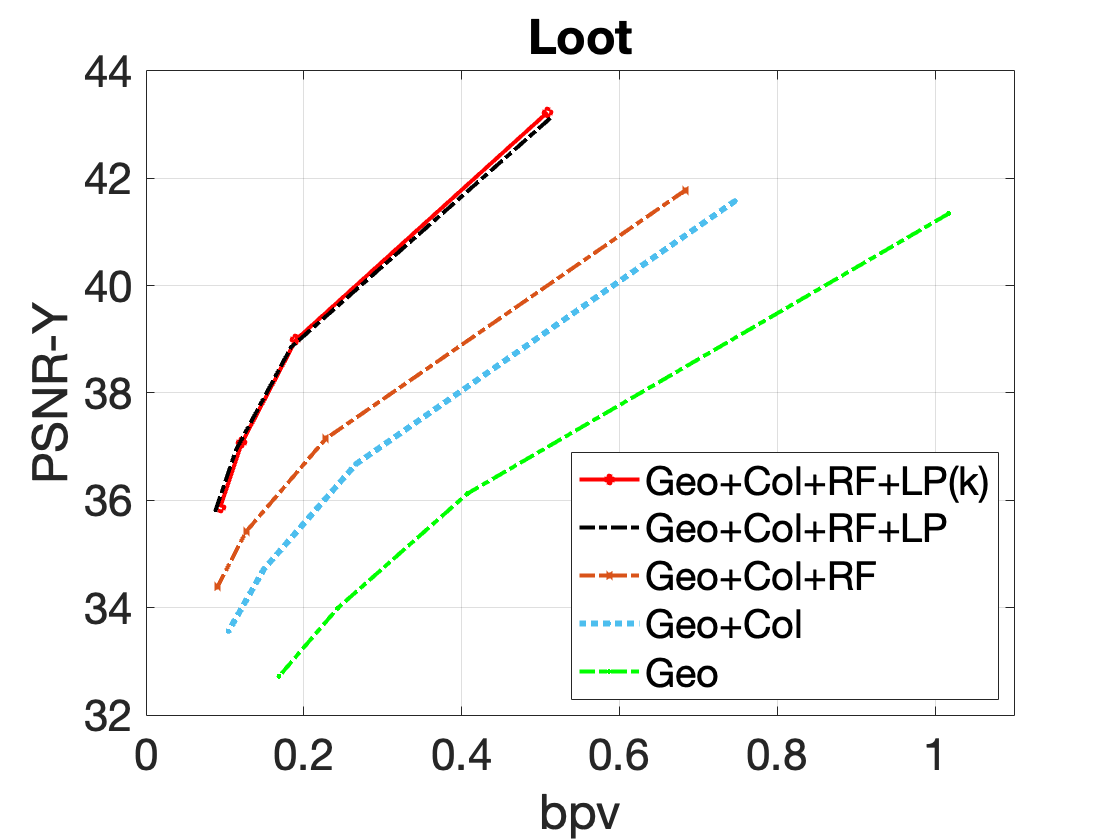}
\end{subfigure}
\caption{Evaluation of proposed ME and in-loop filters.}
\label{fig:RDcurves1}
\vspace{-2mm}
\end{figure}

From \autoref{fig:RDcurves1}, comparing  \texttt{Geo+Col} and \texttt{Geo+Col+RF} to \texttt{Geo} we can see that both the color-ICP  (\autoref{sec:ICPmotion}) and local refined search  (\autoref{sec:RFmotion}) lead to performance improvements. The largest boost in performance is obtained with the low pass filtered prediction of \autoref{sec:LP} (see  \texttt{Geo+Col+RF+LP}), while adaptive filtering can provide additional gains for some sequences. 
\subsection{Effect of in-loop filtering}
\begin{figure}[htb]
\begin{subfigure}[t]{0.25\textwidth}
\includegraphics[width = \linewidth]{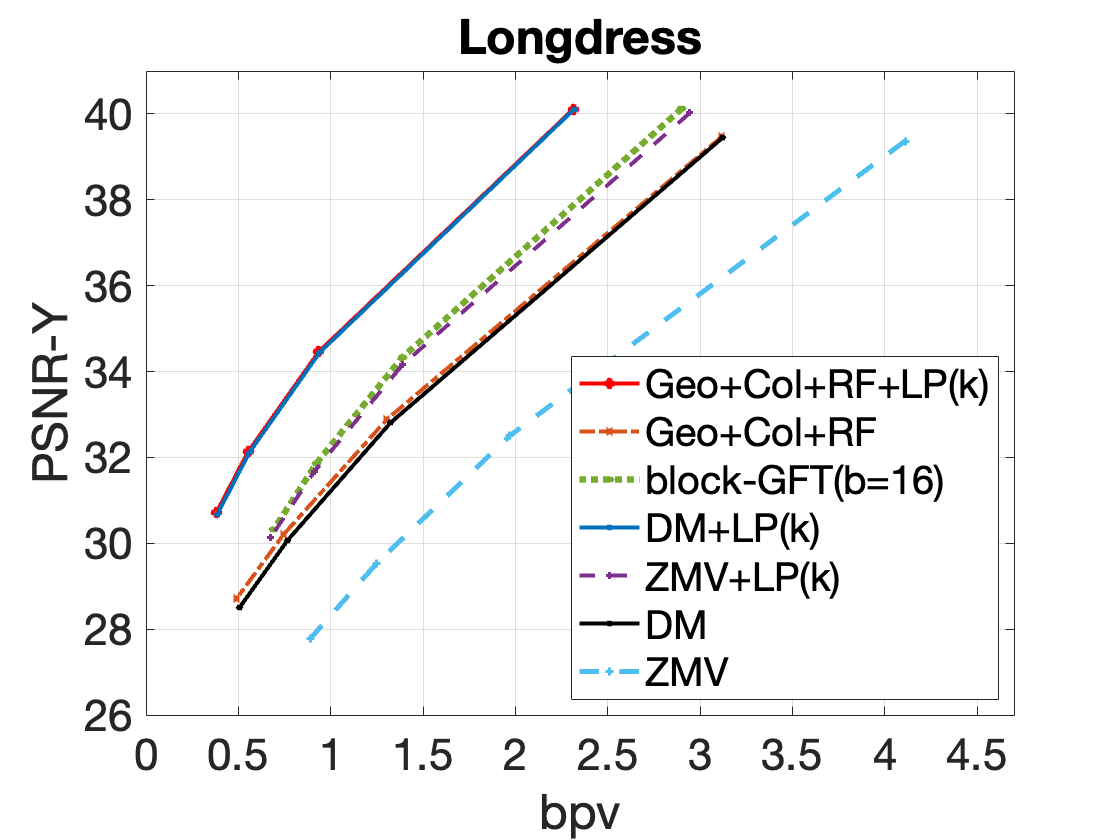}
\end{subfigure}
\hspace{-5.5mm}
\begin{subfigure}[t]{0.25\textwidth}
\includegraphics[width = \linewidth]{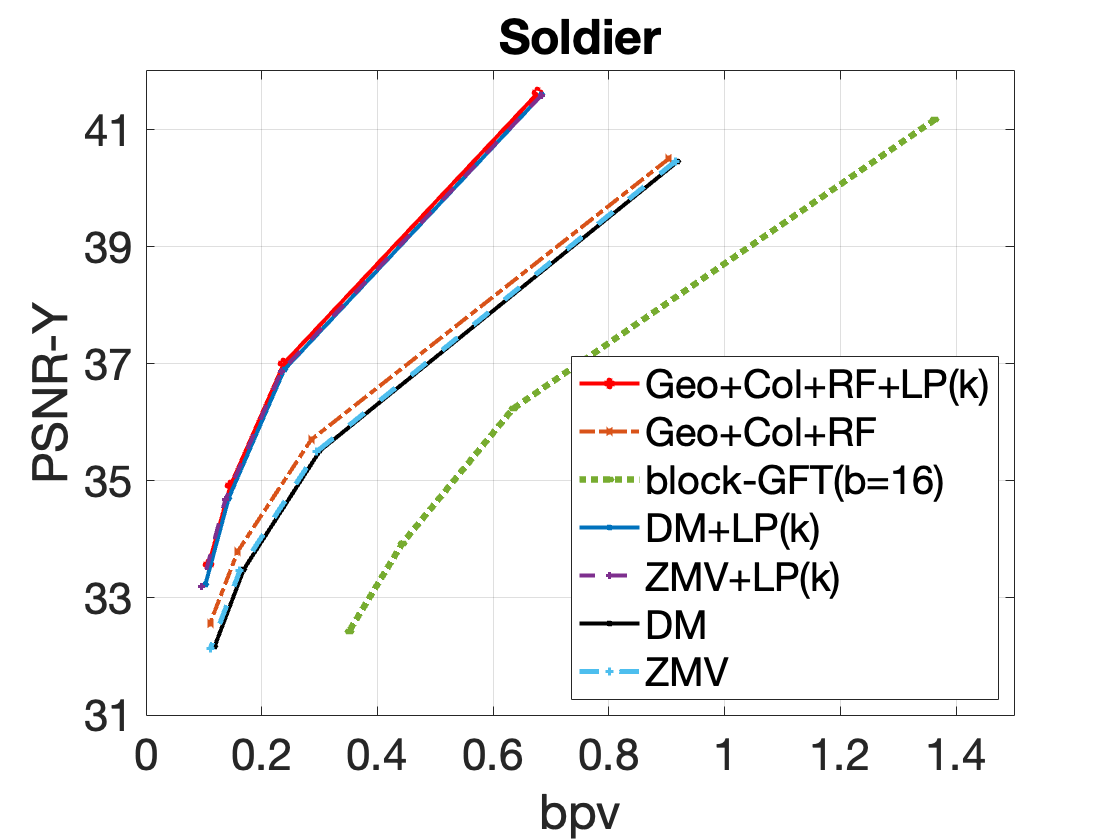}
\end{subfigure}
\hspace{-6mm}
\begin{subfigure}[t]{0.25\textwidth}
\includegraphics[width = \linewidth]{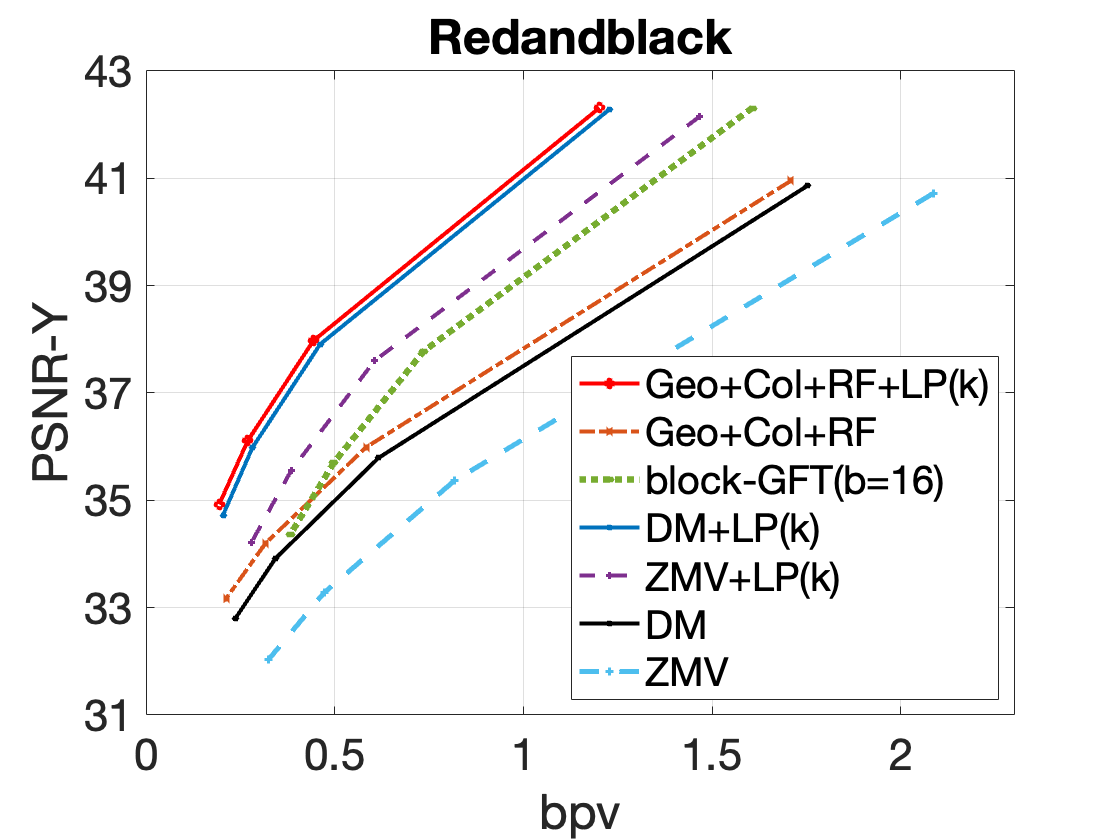}
\end{subfigure}
\hspace{-5mm}
\begin{subfigure}[t]{0.25\textwidth}
\includegraphics[width = \linewidth]{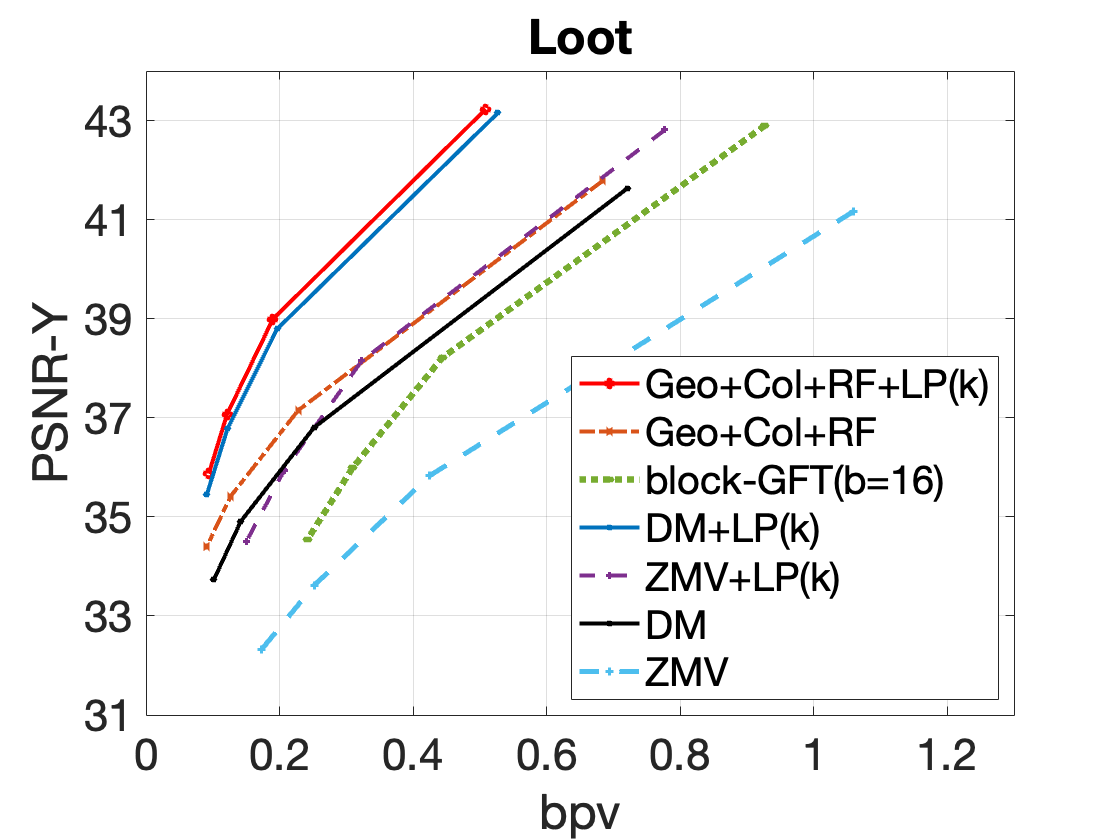}
\vspace{-6mm}
\end{subfigure}
\hspace{-6mm}
\caption{Rate distortion curves of the combination of the proposed filtering scheme and the existing ME schemes.}
\label{fig:RDcurves2}
\vspace{-2mm}
\end{figure}

%
In this section we further evaluate the proposed in-loop filtering approaches  when combined with other ME techniques. We considered zero motion vectors (ZMV) \cite{Souto2020} and the database motion (DM) \cite{IMVdataset} for $16\times16\times16$  blocks, which are both combined with our proposed adaptive filtering, leading to  \texttt{ZMV+LP(k)} and \texttt{DM+LP(k)}. 
To illustrate how the proposed adaptive filtered prediction scheme can improve performance when combined with other methods, including methods that generate inaccurate motion, 
we exclude the mode decision function proposed in \cite{Souto2020}.  
After prediction, the residues are transformed by block-GFT  \cite{GFT2014} with block size equal to 16. The intra coding baseline using block-GFT scheme denoted as \texttt{block-GFT(b=16)} is also included in the experiment. To further demonstrate the performance of our ME scheme, we consider the above schemes without filtering as anchors denoted as \texttt{ZMV} and \texttt{DM} respectively. Our results are shown in \autoref{fig:RDcurves2}.


Combined with the proposed adaptive low pass filtering scheme, both \texttt{ZMV} and \texttt{DM} schemes are improved significantly. \texttt{ZMV+LP(k)} can even outperform intra coding method, \texttt{block-GFT (b=16)}, in some sequences without using mode decision to select between intra and inter coding. 
This is because our adaptive filtering scheme can aggressively eliminate AC components of predictors while preserving DC components. When the available motion is inaccurate, a filtered prediction with large $k$ is more likely to be selected. 
In this case, MCP will mostly eliminate DC components of predicted signals while AC components remain unchanged. 
\texttt{ZMV+LP(k)} can be viewed as being reduced to transform scheme \texttt{RA-GFT}. 
From this perspective, our adaptive filtered prediction scheme plays a role similar to mode decision for \texttt{ZMV}. 
The combination of our efficient filtering scheme and ZMV, \texttt{ZMV+LP(k)}, can be a promising scheme for real-time applications, since it avoids high computational cost ME and mode decision. Due to the local refined search, \texttt{Geo+Col+RF} can always outperform \texttt{DM}, which illustrates that the motion computed by our proposed ME scheme is more accurate than \texttt{DM} for color prediction. 

\subsection{Comparison with state-of-the-art methods}
We compare our method against our implementation of the patch-based inter-coding scheme of \cite{Xu2020} denoted as \texttt{predictive NWGFT}, which includes K-means clustering, geometric ICP-based ME, optimal inter-prediction, predictive transform 
and GFT based on the normal-weight graph (NWGFT). 
As discussed previously, for simplicity, none of our implementations, including that of \cite{Xu2020}, makes use of RD optimized mode decision. We leave this for future work.  
We use the same parameter settings in the paper \cite{Xu2020} except the $\epsilon^2$ in NWGFT, which we set to $\epsilon^2=3$ instead of $\epsilon^2=50$\footnote{We observe that   a smaller $\epsilon$ leads to significantly better performance.}. 
Since the ME in \cite{Xu2020} only depends on the geometry, no overhead is required to transmit the MVs. 
Additionally, to make our performance evaluation more comprehensive, we include two state-of-the-art non-predictive anchor solutions, namely, the region adaptive graph Fourier transform (\texttt{RA-GFT}) with block size $16$ \cite{RAGFT},  and the region adaptive Haar transform (\texttt{RAHT}) \cite{raht}.  

\begin{figure}[htb]
\begin{subfigure}[t]{0.25\textwidth}
\includegraphics[width = \linewidth]{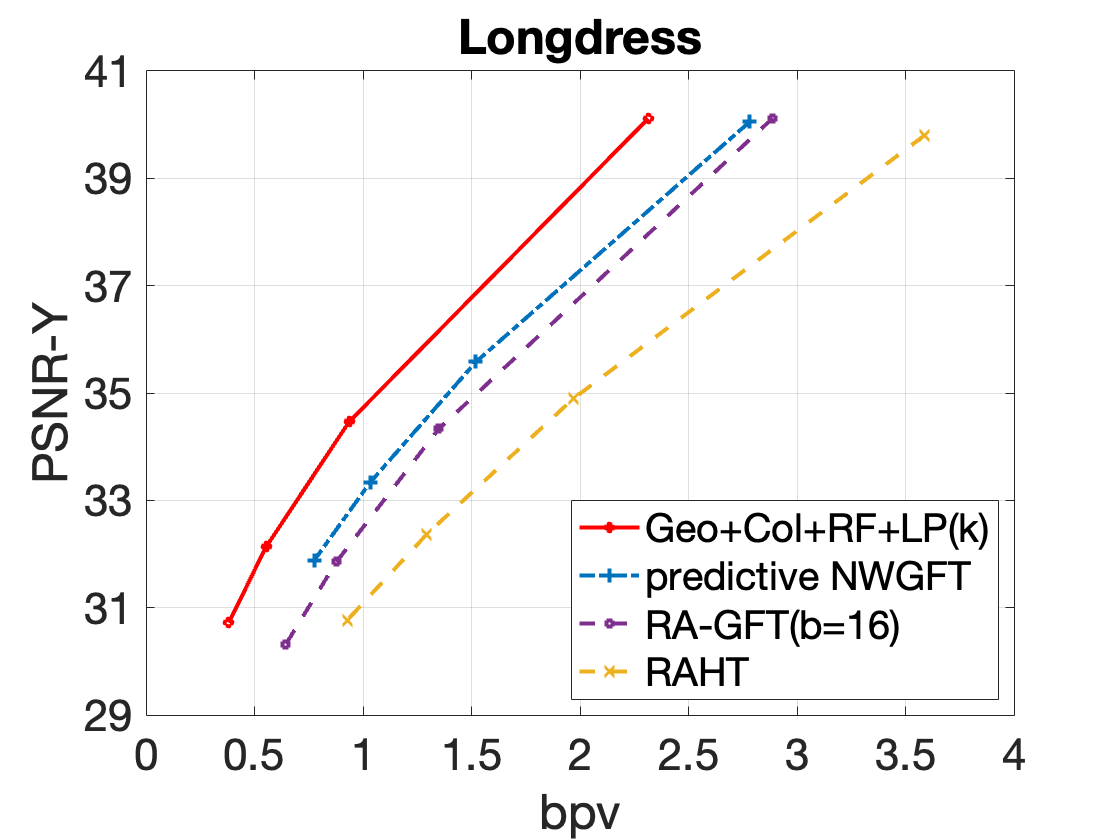}
\vspace{-6mm}
\end{subfigure}
\hspace{-6mm}
\begin{subfigure}[t]{0.25\textwidth}
\includegraphics[width = \linewidth]{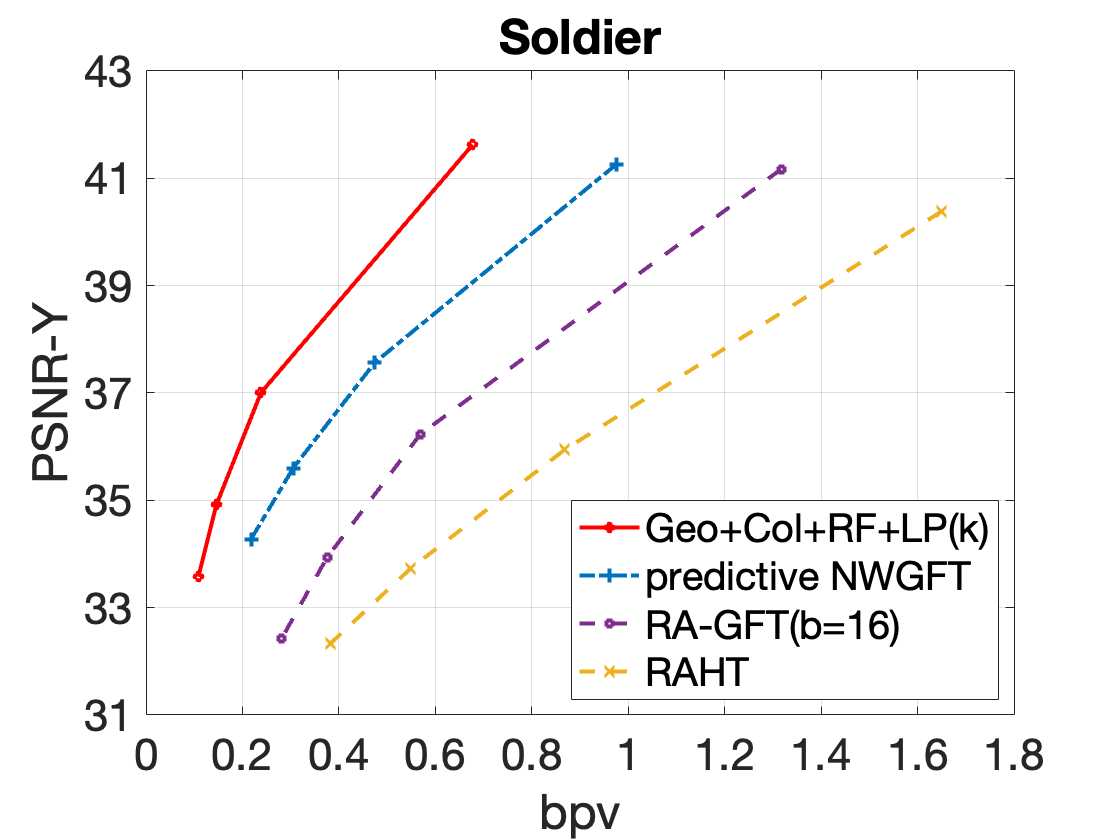}
\end{subfigure}
\hspace{-6.1mm}
\begin{subfigure}[t]{0.25\textwidth}
\includegraphics[width = \linewidth]{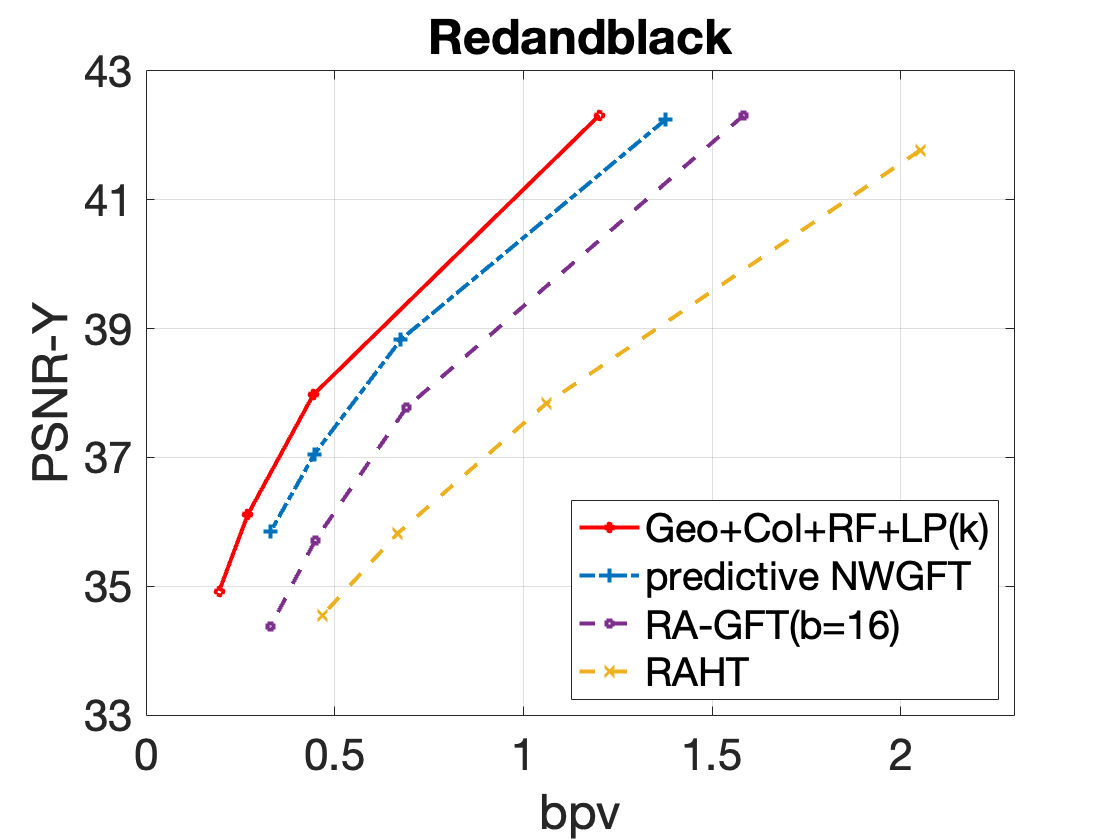}
\end{subfigure}
\hspace{-5mm}
\begin{subfigure}[t]{0.25\textwidth}
\includegraphics[width = \linewidth]{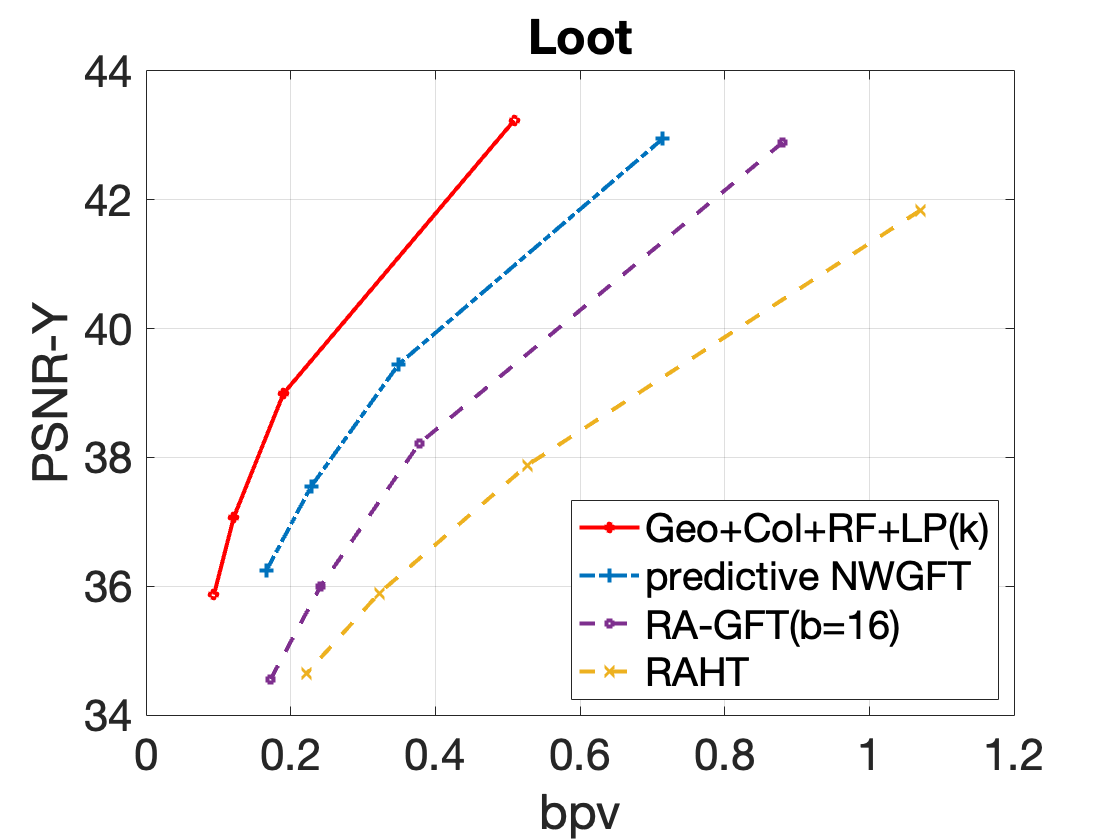}
\end{subfigure}
\hspace{-6mm}
\caption{Rate distortion curves of the proposed scheme and state of the art schemes.}
\label{fig:RDcurves3}
\vspace{-2mm}
\end{figure}

We can see  from \autoref{fig:RDcurves3} that 
our inter coding scheme can consistently and significantly outperform transform schemes such as \texttt{RA-GFT(b=16)} and \texttt{RAHT}, with average gains of $3.21dB$ and $4.95dB$, respectively, even without using a mode decision function or a rate-distortion optimization technique. 
We attribute our superior performance to an effective estimation of motion to exploit temporal redundancy. 
Our method can also outperform competitive \texttt{predictive NWGFT} with average gains of $2.15dB$ because of the more accurate motion by the integration of color attributes during ME and the proposed filtered prediction scheme.

\section{conclusion}
\label{sec:conclusion}
In this paper, we propose an inter predictive approach  for compression of  dynamic point cloud attributes. The proposed encoder consists of an efficient motion estimation scheme and an in-loop  low-pass filtering scheme for improved motion compensated prediction. 
Our motion estimation scheme uses color-ICP   combined with a refined local search scheme to search for the motion that minimizes color error. We also propose an efficient vertex domain  graph-based low-pass filter to  remove noise from reconstructed reference signals. The  proposed filtered prediction scheme that combines the aforementioned techniques, 
significantly outperforms  state of the art approaches for  attribute compression in dynamic 3D point clouds.

\bibliographystyle{ieeetr}
\bibliography{refs.bib}

\end{document}